# Multifaceted Exploration of Spatial Openness in Rental Housing

*A Big Data Analysis in Tokyo's 23 Wards*

**Takuya Oki and Yuan Liu**
*Institute of Science Tokyo, School of Environment and Society, Japan.*
*oki.t.e60f@m.isct.ac.jp, ORCID: 0000-0002-4848-0707*

**Abstract.** Understanding spatial openness is vital for improving residential quality and design; however, studies often treat its influencing factors separately. This study developed a quantitative framework to evaluate the spatial openness in housing from two- (2D) and three- (3D) dimensional perspectives. Using data from 4,004 rental units in Tokyo's 23 wards, we examined the temporal and spatial variations in openness and its relationship with rent and housing attributes. 2D openness was computed via planar visibility using visibility graph analysis (VGA) from floor plans, whereas 3D openness was derived from interior images analysed using Mask2Former, a semantic segmentation model that identifies walls, ceilings, floors, and windows. The results showed an increase in living room visibility and a 1990s peak in overall openness. Spatial analyses revealed partial correlations among openness, rent, and building characteristics, reflecting urban redevelopment trends. Although the 2D and 3D openness indicators were not directly correlated, higher openness tended to correspond to higher rent. The impression scores predicted by the existing models were only weakly related to openness, suggesting that the interior design and furniture more strongly shape perceived space. This study offers a new multidimensional data-driven framework for quantifying residential spatial openness and linking it with urban and market dynamics.

**Keywords.** Spatial openness, Rental housing, Floor plan, Interior image, Computer vision.

## 1. Introduction

Understanding spatial openness in residential housing is important for improving quality of life and architectural design. Several factors influence spatial openness. However, few studies have attempted to evaluate spatial openness quantitatively while considering such diverse factors.

The study of spatial openness has been approached from various perspectives in architectural and environmental psychology. Traditional spatial analysis methods focus on geometric properties and numerical metrics of building attributes. The space syntax theory (Hillier et al., 1987) introduced systematic approaches to understanding spatial relationships through a graph-based analysis of architectural layouts. Visibility graph analysis (VGA), which is an extension of space syntax principles, provides a quantitative framework for measuring visual connectivity and spatial integration within architectural spaces. Recent studies have expanded the understanding of spatial openness to include cultural and contextual dimensions. Al-Mohannadi et al. (2020) demonstrated how spatial configurations in traditional Qatari domestic architecture embody the cultural values of privacy and openness through sophisticated spatial arrangements, highlighting the importance of considering cultural contexts in spatial analysis. This perspective emphasises that spatial openness is not merely a physical property but a culturally mediated experience that varies across different architectural traditions. Methodological approaches for analysing spatial configurations have also evolved significantly. Kim and Choi (2009) introduced angular and cellular VGA as innovative extensions of traditional space syntax methods, providing more nuanced approaches to analyse human movement behaviour in architectural spaces. These methodological advances enable researchers to capture the dynamic aspects of the spatial experience that conventional VGA may overlook.

Previous studies demonstrated that spatial openness influences occupant behaviour, movement patterns, and psychological responses to built environments. Environmental psychology studies have shown that perceived openness is correlated with reduced stress levels, improved cognitive performance, and enhanced overall well-being. However, most existing research is limited to small-scale case studies or controlled experimental settings, lacking the scale and diversity necessary to understand spatial openness patterns across all urban regions.

Recent advances in computer vision and deep learning have opened up new possibilities for automated spatial analysis. Semantic segmentation techniques enable the extraction of architectural elements from images, allowing for the systematic analysis of interior compositions. Additionally, the emergence of big data analytics in housing research has revealed new patterns in residential preferences, market dynamics, and architectural trends. Online rental platforms generate vast amounts of data including property specifications, pricing information, and visual documentation through interior images and floor plans. This data richness enables researchers to investigate spatial quality and market valuation at unprecedented scales, creating opportunities that were previously impossible because of data limitations. For instance, there have explorations of retrieving more abstract representations of data types, such as floor plans, using machine-driven automated approaches (Kitabayashi et al., 2022); however, these methods tend to focus on structural representations. Additionally, while research focusing on preferences regarding floor plans and interior design does exist, analyses focusing on three-dimensional (3D) spatial composition and architectural elements are lacking (Oki and Shimomura, 2025).

In this study, we developed a method for quantitatively evaluating the spatial openness of residential buildings from multiple perspectives using big data on rental housing in Tokyo's 23 wards. The, we analysed regional and temporal variations in the spatial openness of rental housing, as well as its impact on rent, based on multiple indicators.

## 2. Methodology

We considered two-dimensional (2D) openness indicators based on floor plans, and

# MULTIFACETED EXPLORATION OF SPATIAL OPENNESS IN RENTAL HOUSING

3D openness indicators based on interior images. Each indicator was calculated from images included in the LIFULL HOME'S dataset with information such as construction year, location, and property area.

## 2.1. DATASET

Based on image quality and data completeness, 4,004 rental properties (0.52%) located in Tokyo's 23 wards that were built after 1960 were extracted (Figure 1). Most properties were excluded because they did not contain an effective living room image, which was necessary for calculating 3D openness.

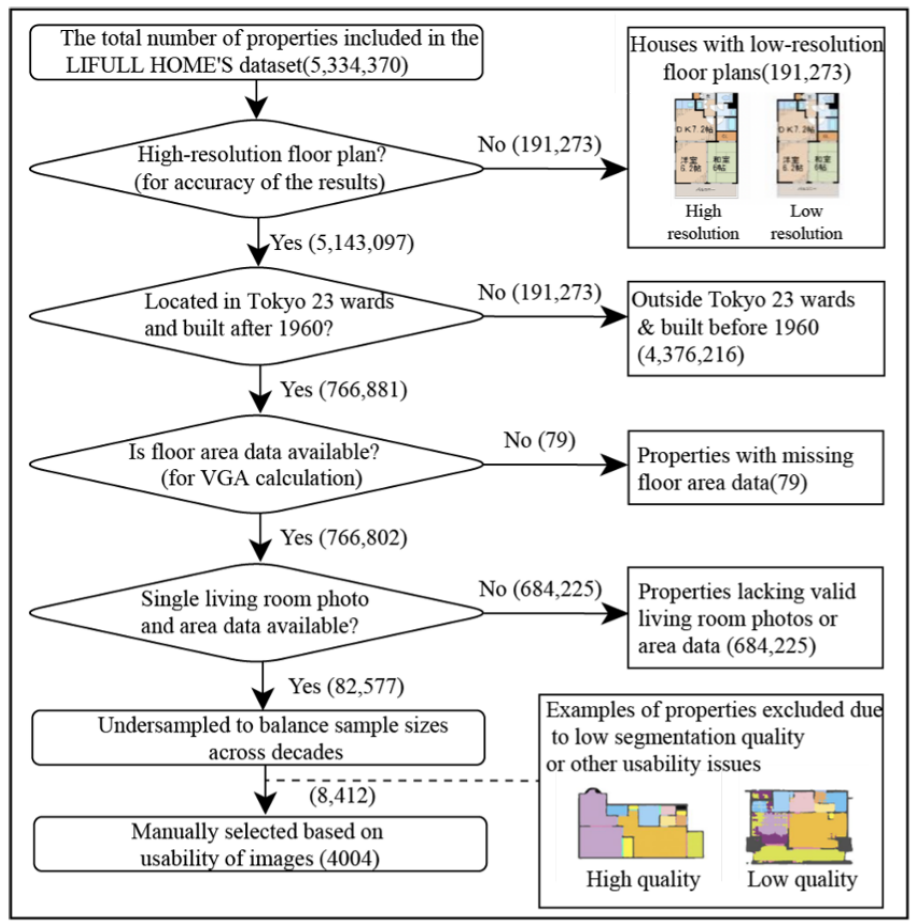

*Figure 1. Data filtering and preprocessing pipeline.*

## 2.2. 2D SPATIAL OPENNESS INDICATORS

We constructed a custom pipeline based on visibility graph analysis (Turner et al., 2001). Our method simulates the inter-visibility among points within a space, enabling automated computation of 2D spatial openness across a large-scale housing dataset without CAD data (Figure 2).

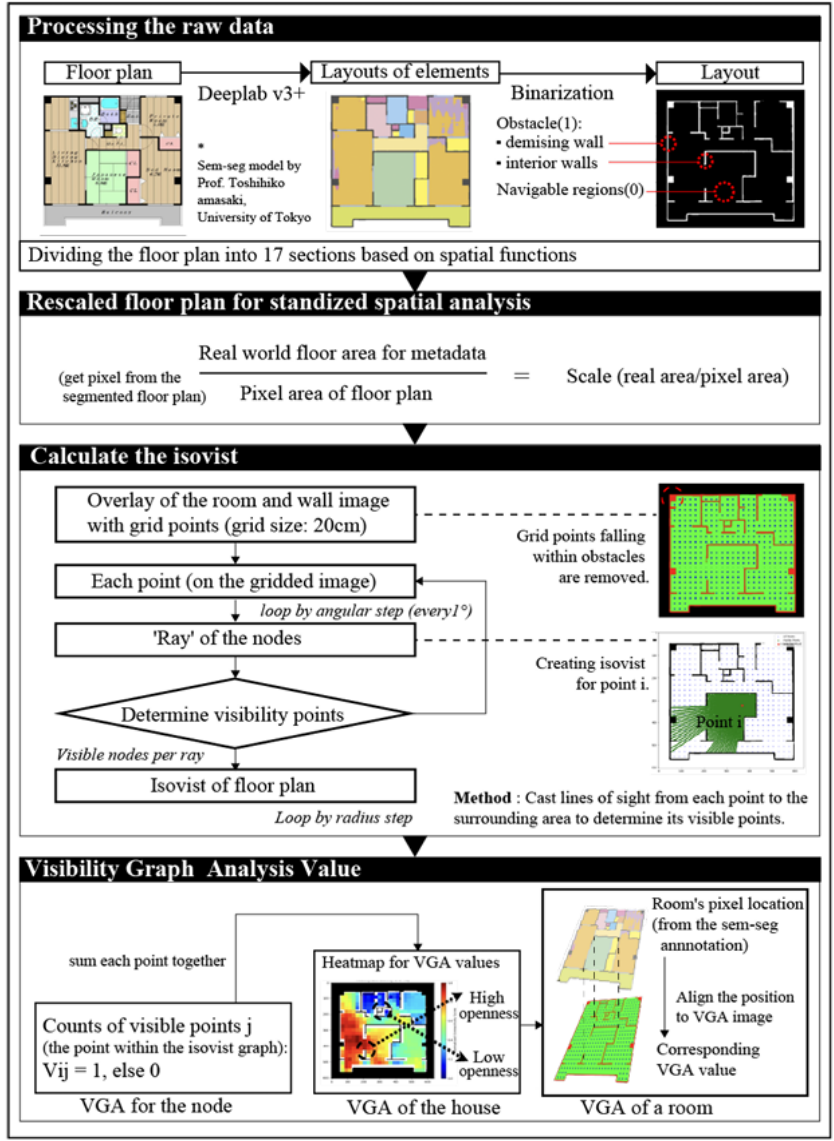

*Figure 2. Computational workflow of 2D spatial openness.*

(1) The original 2D floor plan images were fed into the DeepLab-V3+ model (Chen et al., 2018), which was fine-tuned for floor plan images to classify pixels into categories such as walls, rooms, and windows (Yamada et al., 2021).

(2) The pixels belonging to the outer boundaries and internal walls were set to 1, whereas the others were set to 0.

(3) All floor plans were rescaled using a factor derived from the actual floor area of each property and the number of pixels in the image.

(4) A uniform grid (20-cm interval) was overlaid on each floor plan.

(5) For each grid point, the total number of grid points that could be connected in a straight line without wall obstructions was counted to construct a visibility graph.

(6) Summary statistics from the visibility graphs were extracted as 2D openness indicators.

## 2.3. 3D SPATIAL OPENNESS INDICATORS

We applied the semantic segmentation task using Mask2Former, pre-trained using the ADE20K dataset (Zhou et al., 2017), to extract the wall, ceiling, floor, and window area ratios from the interior images. The fractions of these visual elements were computed by dividing the number of pixels corresponding to each element by the total number of visible pixels in the image (Figure 3).

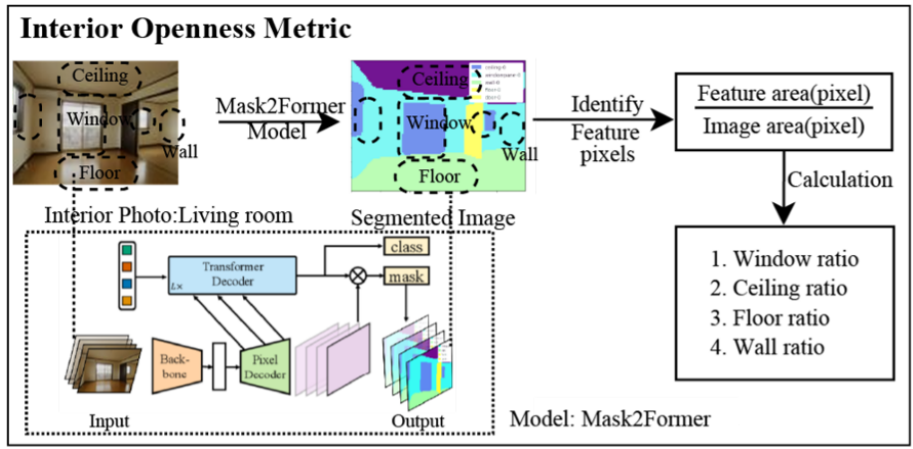

*Figure 3. Computational workflow of 3D spatial openness.*

## 3. Results

### 3.1. EXAMPLES OF VISUALISING 2D/3D OPENNESS

Figure 4 shows an example of the visualised visibility maps and segmentation results when calculating the actual 2D/3D openness from floor plans and interior images included in the LIFULL HOME'S dataset.

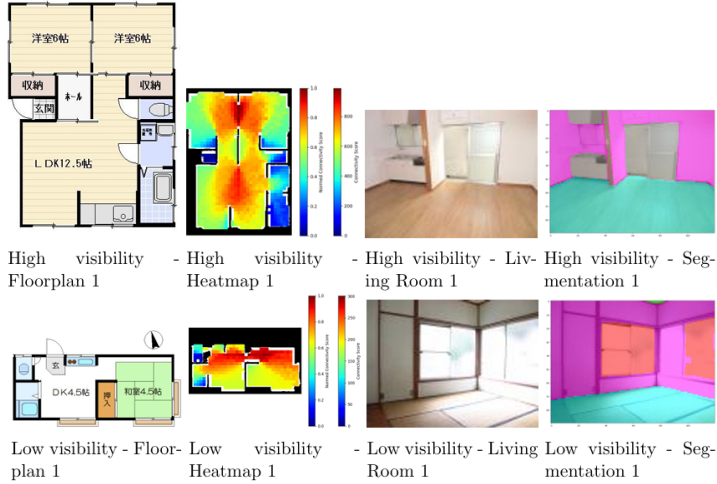

High visibility - Floorplan 1 | High visibility - Heatmap 1 | High visibility - Living Room 1 | High visibility - Segmentation 1

Low visibility - Floorplan 1 | Low visibility - Heatmap 1 | Low visibility - Living Room 1 | Low visibility - Segmentation 1

*Figure 4. Examples of visualizing 2D/3D openness for property with high or low visibility.*

### 3.2. TEMPORAL TREND OF OPENNESS INDICATORS

The 2D visibility scores of the entire house peaked in the 1990s and declined thereafter, indicating that the openness of the overall floor plans remained relatively stable in recent decades. The standard deviation increased over time, indicating that rental housing is becoming more complex in terms of 2D openness. From the perspective of 3D openness, over the years, the proportion of ceilings in the images tended to increase, whereas the proportion of windows tended to decrease (Figure 5).

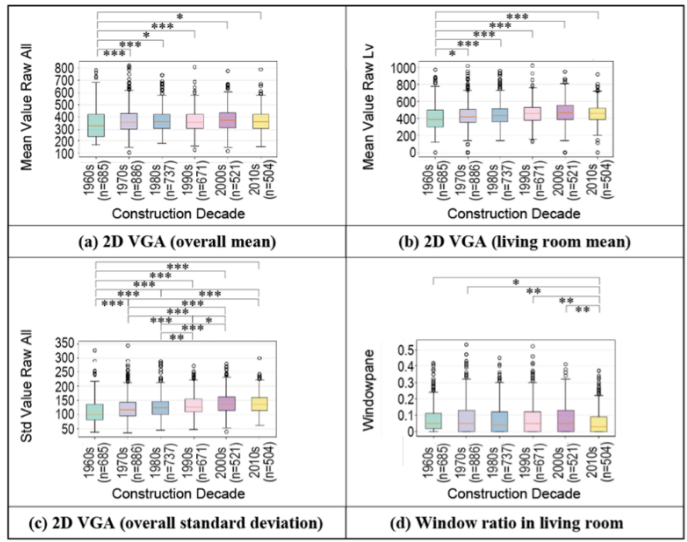

*Figure 5. Temporal trends of openness indicators (\*\*\*: 0.1%, \*\*: 1%, \*:5% significance).*

### 3.3. GEOSPATIAL DISTRIBUTION OF OPENNESS INDICATORS

The spatial patterns of openness, rent, construction year, and floor area were partially correlated, indicating that urban redevelopment and housing design trends were manifested geographically (Figure 6). The ratio of living-room windows tended to be high in the city centre. This may have been influenced by the existence of high-rise apartments, in which views are valued. The relatively low 2D openness values in suburban areas are believed to be influenced by housing types with many individual rooms and housing complexes.

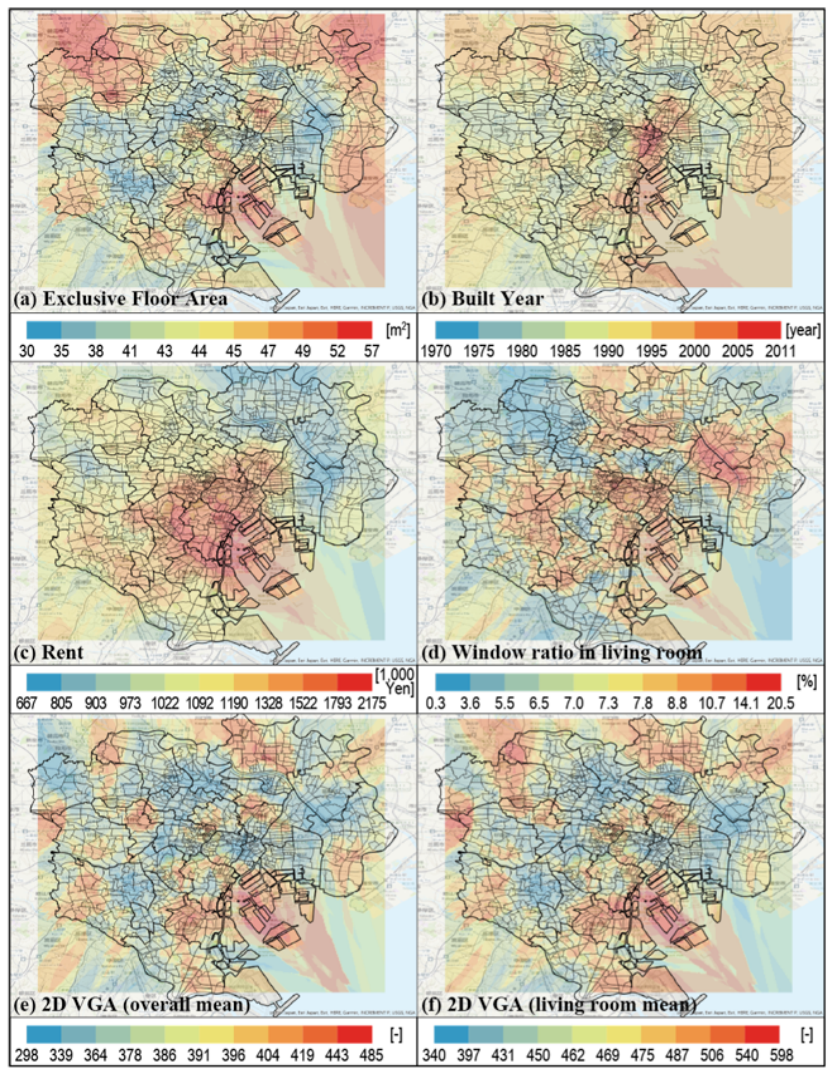

*Figure 6. Geospatial distribution of openness indicators.*

### 3.4. CORRELATION ANALYSIS AMONG INDICATORS AND WITH RENT AND OTHERS

No clear correlation was found between the 2D and 3D openness indicators; however, a high correlation was confirmed between the mean value and standard deviation of 2D openness. The 3D openness indicators were generally complementary. Additionally, there was a tendency for properties with higher openness to have higher rent (Figure 7).

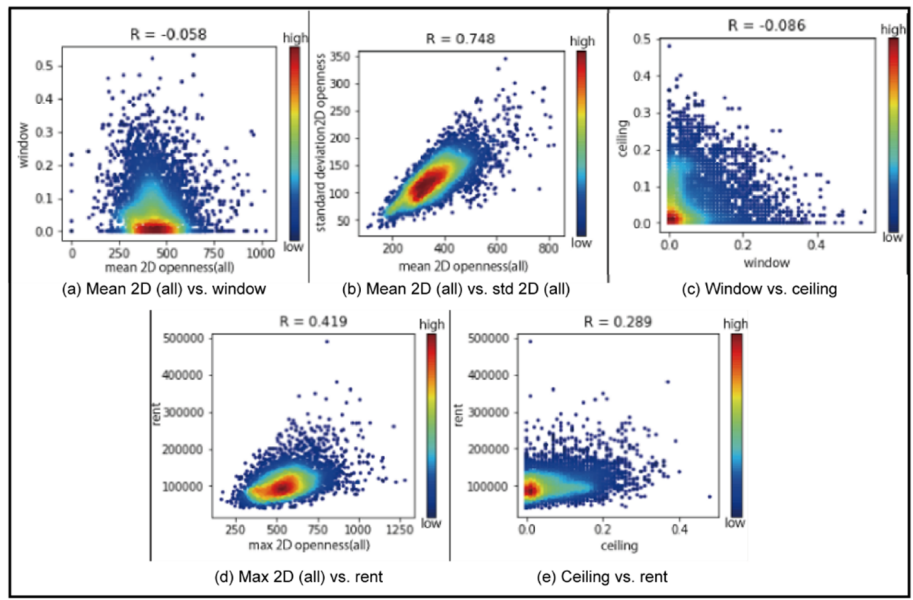

*Figure 7. Correlation analysis among indicators and with rent and others.*

### 4. Discussion

Although this study advances the quantification and analysis of spatial openness in residential environments, limitations and areas for improvement were identified that warrant further investigation.

One important limitation of the 3D openness features in this study is the lack of standardisation of image data. The interior images were captured from various viewpoints with different camera angles, heights, and depths, all of which could significantly affect the resulting 3D feature metrics. Particularly, houses photographed from extremely high or low angles exhibit biased representations of architectural features. Even when the spatial structure remains constant, differences in shooting parameters may cause substantial variations in the extracted indicators, undermining their stability, interpretability, and comparability across samples.

Additionally, the dataset of 4,004 properties, while substantial for a study involving detailed spatial analysis, represents a limited sample, given Tokyo's vast rental market. The filtering process necessary to ensure the data quality further constrained the sample size. Future research should address the expansion of larger datasets by improving the automation of data-processing pipelines.

## 5. Conclusion

We developed indicators to quantitatively evaluate spatial openness from multiple perspectives, such as horizontal and elevational, using big data on rental housing in Tokyo's 23 wards. Then, we analysed the changes in spatial openness over time, geospatial characteristics, and correlations between indicators.

Although this study focused on Tokyo's 23 wards, the proposed quantitative evaluation method for spatial openness is robust. As long as comparable floor plans and interior images are available, this method can be applied to housing studies in other cities. This framework holds strong potential for real-world applications: it can support architects and planners in optimising spatial layouts to enhance openness, especially under constraints of limited space and budget; can be integrated into housing recommendation systems to provide more personalised suggestions based on openness preferences; and can also inform urban planning and policymaking by linking openness data with demographic changes and housing supply trends. These future directions not only address the current limitations but also broaden the impact of spatial openness research, ultimately contributing to the creation of better-designed living environments.

### Acknowledgements

In this study, we used the LIFULL HOME'S Dataset provided by LIFULL Co., Ltd. via the IDR Dataset Service of the National Institute of Informatics. We would like to thank Professor Toshihiko Yamasaki (University of Tokyo) for sharing the program used to perform semantic segmentation.

<Attribution with AI Tools>
Grammarly (Grammarly Inc., 2024) was used to correct errors in spelling and grammar.